\documentclass{article}

    \usepackage[preprint]{neurips_2025}

\usepackage[utf8]{inputenc} % allow utf-8 input
\usepackage[T1]{fontenc}    % use 8-bit T1 fonts
\usepackage{hyperref}       % hyperlinks
\usepackage{url}            % simple URL typesetting
\usepackage{booktabs}       % professional-quality tables
\usepackage{amsfonts}       % blackboard math symbols
\usepackage{nicefrac}       % compact symbols for 1/2, etc.
\usepackage{microtype}      % microtypography
\usepackage{xcolor}         % colors
\usepackage{multirow}
\usepackage{booktabs}
\usepackage{colortbl}       % color for table rows
\usepackage{amsmath,amssymb}
\usepackage{amsthm}
\usepackage{subcaption}
\usepackage{graphicx}
\usepackage{wrapfig}
\newtheorem{theorem}{Theorem}[section]

\title{Uncertainty-quantified Rollout Policy Adaptation for Unlabelled Cross-domain Temporal Grounding}

\author{%
  Jian Hu\textsuperscript{1}, Zixu Cheng\textsuperscript{1} Shaogang Gong\textsuperscript{1}, Isabel Guan\textsuperscript{2}, Jianye Hao\textsuperscript{3}
, Jun Wang\textsuperscript{4}
, Kun Shao\textsuperscript{3}\thanks{Corresponding author: shaokun2@huawei.com}  \\
  \textsuperscript{1}Queen Mary University of London, \textsuperscript{2}Hong Kong University of Science and Technology, \\\textsuperscript{3}Huawei Noah’s Ark Lab,  \textsuperscript{4}University College London 
}

\begin{document}

\maketitle

% \vspace{-10pt}
\begin{abstract}
Video Temporal Grounding (TG) aims to temporally locate video segments matching a natural language description (a query) in a long video. While Vision-Language Models (VLMs) are effective at holistic semantic
matching, they often struggle with fine-grained temporal
localisation. Recently, Group Relative Policy Optimisation (GRPO)
reformulates the inference process as a reinforcement learning
task, enabling fine-grained grounding and achieving strong in-domain
performance. However, GRPO relies on labelled data, making it unsuitable in unlabelled domains. Moreover, because videos are large and expensive to store and process, performing full-scale adaptation introduces prohibitive latency and computational overhead, making it impractical for real-time deployment. To
overcome both problems, we
introduce a Data-Efficient Unlabelled Cross-domain Temporal Grounding
method, from which a model is first trained on a labelled source domain, then adapted to a target domain using only a small number of {\em unlabelled videos from the target domain}. This approach eliminates the need for target annotation and keeps both computational and storage overhead low enough to run in real time. Specifically, we introduce
\textbf{U}ncertainty-quantified \textbf{R}ollout \textbf{P}olicy
\textbf{A}daptation (\textbf{URPA}) for cross-domain knowledge transfer in learning video temporal grounding without target labels. URPA generates multiple candidate predictions using
GRPO rollouts, averages them to form a pseudo label, and estimates
confidence from the variance across these rollouts. This confidence
then weights the training rewards, guiding the model to focus on
reliable supervision. Experiments on three datasets across six cross-domain settings show that URPA generalises well using only a few unlabelled target videos. Codes will be released once published.
%
  % Temporal grounding (TG) aims to localise the temporal segment of a video that corresponds to a given natural language query. However, Vision-Language Models (VLMs) often prioritise global semantic matching over fine-grained temporal alignment, limiting their effectiveness in this task.
  % %
  % Recently, GRPO employs a metric-based optimisation objective, guiding the model to capture fine-grained details instead of relying solely on global semantic matching.
  % %
  % Inspired by this, we propose a data-efficient cross-domain temporal grounding task. Unlike traditional cross-domain settings, we assume after pretraining on labelled source domain, only a small number of samples are available from the unsupervised target domain, enabling the model to generalise effectively with minimal training data. 
  % %
  % Specifically, we first train the model with the labelled source domain to acquire basic grounding ability, and then perform cross-domain adaptation in the data-efficient target domain using our proposed unsupervised GRPO strategy. 
  %
  % Extensive experiments demonstrate the effectiveness and generalisability of our approach.
  %
  
\end{abstract}

\section{Introduction}
Temporal Grounding (TG) localises the exact temporal segment in an untrimmed video that semantically corresponds to a given natural-language query~\cite{anne2017localizing,gao2017tall}. Accurate TG is fundamental to high-level applications such as activity detection~\cite{dong2022partially} and embodied human–computer interaction~\cite{dong2021dual}. 

\begin{figure*}[t]
  \centering
  \vspace{-7pt}
  \begin{tabular}{cc} 
\hspace{-8mm}\includegraphics[width=0.345\columnwidth]{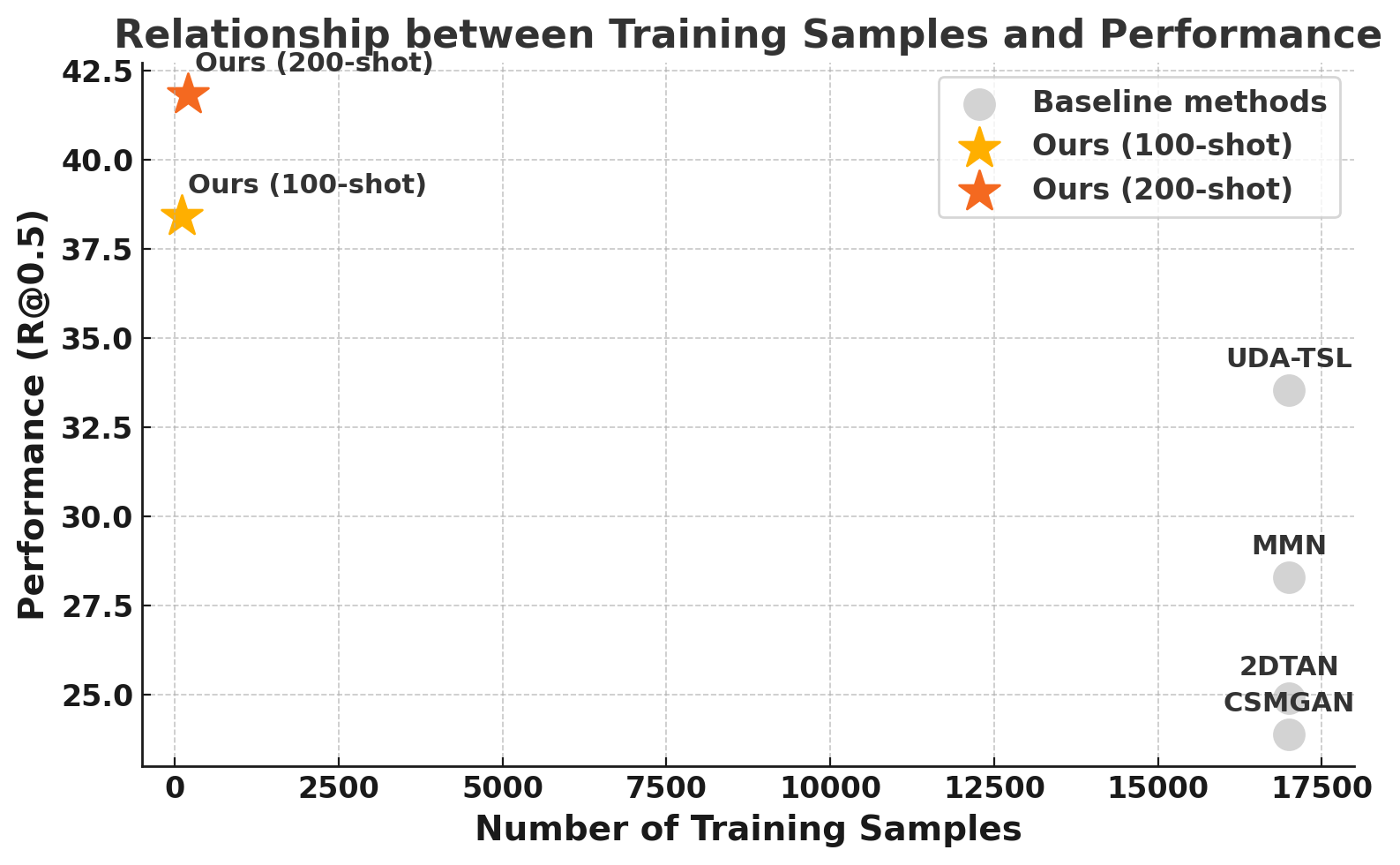} & \hspace{-11mm}\includegraphics[width=0.68\columnwidth]{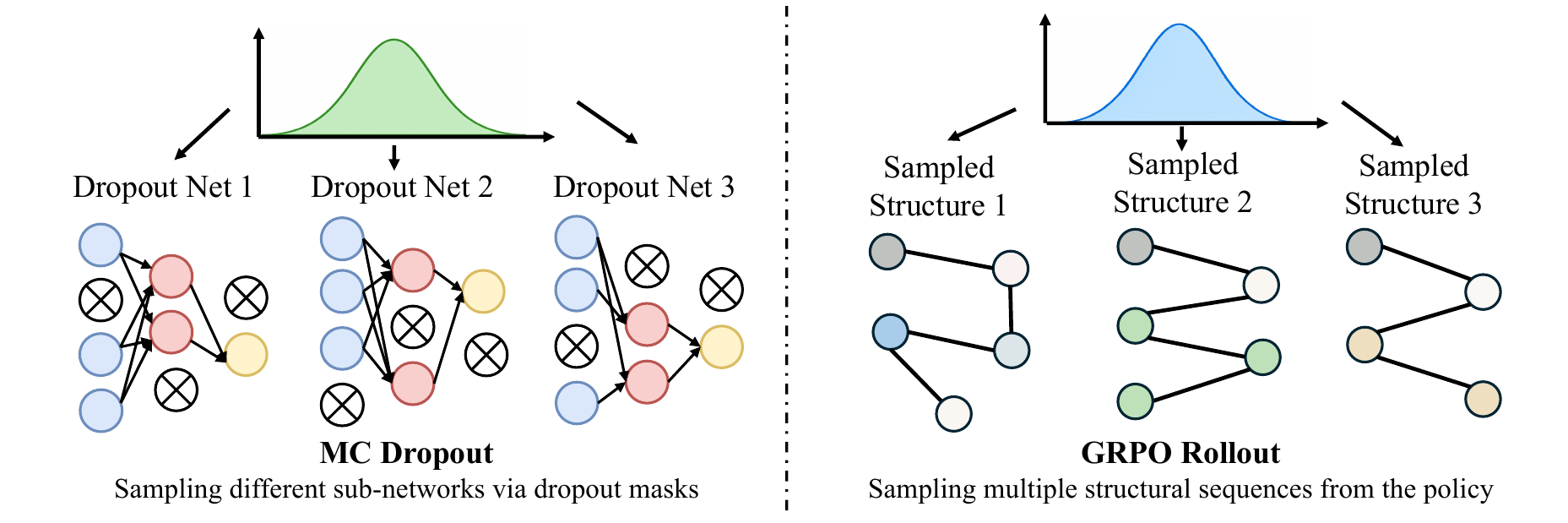}\vspace{-2pt}
\\
    \hspace{-3mm}  
    {\small{\selectfont(a) Motivation of data-efficient CTG setting.}} & \hspace{-5mm}  
    {\small{\selectfont(b) Comparison between MC Dropout and GRPO rollout.}} 

  \end{tabular}
  % \vskip -0.15cm
   \caption{
    (a) A comparison between full-data and data-efficient adaptation in Cross-domain Temporal Grounding (CTG). Existing methods adapt on thousands of unlabelled target videos (grey), which is slow and resource-heavy. We propose a data-efficient CTG setting using only 100 or 200 randomly selected target videos. Despite the limited data, our method matches or exceeds performance on the TaCoS $\rightarrow$ ActivityNet task.
   (b) Conceptual comparison between MC Dropout and GRPO rollout. MC Dropout samples subnetworks via stochastic neuron dropout and estimates uncertainty from output diversity. GRPO rollouts similarly sample diverse structural sequences from the policy. URPA leverages this property to generate averaged pseudo labels and estimate uncertainty via rollout standard deviation, enabling uncertainty-quantified adaptation without ground-truth labels. 
   %See Sec.~\ref{sec:theory} for theoretical analysis of rollout variance-based uncertainty.
   }
  % \label{fig:biases}  
  \label{fig:moti}
  \vspace{-15pt}
\end{figure*}

While VLMs are effective at capturing holistic video semantics~\cite{jin2024chat,tang2025video,hu2025cos}, they often struggle with fine-grained localisation, leading to suboptimal performance in temporal grounding.
Some recent works apply Supervised Fine-Tuning (SFT) to better align video-query pairs~\cite{chung2024scaling,ren2024timechat,zeng2024timesuite}. However, since relevant segments typically cover only a small portion of the video, the model often overfits to redundant context, limiting its ability to perform precise grounding~\cite{zhang2023no}.
To improve temporal reasoning, Chain-of-Thought (CoT) post-training introduces explicit intermediate reasoning steps before prediction~\cite{qin2024grounding, xiong2024large}. While effective, this approach depends on manually annotated prompts, which are expensive to collect and hard to scale across domains and tasks.
More recently, GRPO~\cite{guo2025deepseek} formulates temporal grounding as a policy learning problem, where reinforcement learning is used to directly optimise segment predictions~\cite{wang2025timezero,li2025temporalr1}. This removes the need for hand-crafted prompts used in post-processing and achieves strong results on in-domain benchmarks. However, GRPO relies on ground-truth temporal boundaries to compute reward signals. requiring extensively labelled training videos. It limits its practical usefulness in real-world scenarios where annotations are unavailable. Moreover, GRPO-based temporal grounding performance drops significantly across domains due to distribution shift, revealing its poor generalisation to unseen data.
In addition to annotation constraints, the sheer scale of video data poses practical challenges. A target domain often contains thousands of videos, which are expensive to store, and time-consuming to adapt. As shown in Fig.~\ref{fig:moti}(a), existing adaptation methods require full retraining on the entire target set. 
Such latency and resource-hungry makes them unsuitable for time-sensitive applications like online or on-device deployment. 
We consider a more practical and generalisable approach to cross-domain temporal grounding where a model trained on a labelled source domain can be efficiently adapted to an unlabelled target
domain using only sparse unlabelled target videos, with minimal latency and resource demands.

To address these challenges, we introduce a Data-Efficient Unlabelled Cross-domain Temporal Grounding approach. A model trained on a labelled source domain is adapted at test-time in deployment using only K unlabelled videos from the target domain ($K\!=\!100\sim200$ in our experiments), allowing for real-time adaptation on limited compute resources. 
The main challenge in this approach lies in effectively leveraging limited unlabelled target data. Without annotated temporal boundaries,
pseudo-labels are noisy and highly uncertain, resulting in poor cross-domain performance. To mitigate this, we introduce explicit uncertainty quantification to assess the reliability of pseudo-labels. 
As illustrated in Fig.\ref{fig:moti}(b), MC
Dropout~\cite{gal2016dropout} estimates uncertainty by sampling multiple sub-networks via stochastic dropout and computing the variance across their outputs. Similarly, GRPO rollouts produce
diverse predictions by sampling from a stochastic policy. Inspired by this parallel, we estimate pseudo label confidence by measuring the standard deviation across multiple rollouts per <video, query> pair, and use it
to guide adaptation with uncertainty-quantification weighted model adaptation.

To this end, we propose Uncertainty-quantified Rollout Policy Adaptation (URPA). After supervised training of a GRPO backbone on the source domain, we perform test-time adaptation by generating G stochastic rollouts for each of the K (K=100/200 in our experiments) randomly selected unlabelled videos in the target domain. A pseudo label is constructed by averaging the predicted boundaries across rollouts, and a small margin relaxation is applied to reduce the impact of outliers.
To estimate the reliability of each pseudo label, we follow the Bayesian view of MC Dropout and use the variance across rollouts as an estimate of uncertainty. Samples with lower variance, indicating higher confidence, are given larger weights during a lightweight gradient update. This enables fast and effective adaptation without introducing significant latency. Sec.~\ref{sec:theory} provides a theoretical analysis of how rollout-based variance quantifies uncertainty.

Our contributions are threefold: (i) We introduce an unlabelled cross-domain temporal grounding approach to test-time model adaptation
using only a small number of unlabelled target-domain videos. 
(ii) We propose Uncertainty-quantified Rollout Policy Adaptation (URPA), the first reinforcement learning-based self-learning adaptation method that combines pseudo-label generation with uncertainty-weighted rewards. URPA enables effective GRPO-based temporal grounding across domains {\em without} requiring ground-truth labels to compute reinforcement signals.
(iii) We theoretically show that rollout variance approximates Bayesian predictive variance, quantifying epistemic uncertainty,  and empirically demonstrate that URPA consistently outperforms strong baselines across six cross-domain video temporal grounding benchmarks.

\vspace{-7pt}
\section{Related Works}
\vspace{-5pt}
\textbf{Test-time Adaptation.}
Test-time domain adaptation (TTDA) adapts a pre-trained model to unlabelled test data exhibiting distribution shift, prior to prediction~\cite{wang2020tent, hu2020discriminative, niu2022efficient}.
TTDA approaches can be broadly divided into backward-free and backward-based methods.
Backward-free methods adapt the model on-the-fly during inference, typically by updating batch normalization (BN) statistics without backpropagation. Representative works include DUA~\cite{mirza2022norm}, which applies a running average to BN layers, and DIGA~\cite{wang2023dynamically}, which aligns distributions for semantic segmentation. While efficient per sample, such methods require adaptation for every test input, which may increase inference latency and introduce instability across samples.
In contrast, backward-based methods adapt the model once using target data before inference begins, enabling faster and more stable test-time prediction. These methods often rely on self-supervised objectives like entropy minimization~\cite{wang2020tent, hu2022learning}.
However, video data is large and difficult to store, and performing test-time adaptation on all videos is time-consuming.
Hence, we leverage only a small number of videos from the target domain to perform data-efficient test-time adaptation via backpropagation, and then apply the adapted model to evaluate the full set test video without further updates. This design ensures both adaptability and real-time efficiency, making it well-suited for practical scenarios with limited target domain data.

\textbf{Temporal Grounding.}
Temporal grounding~\cite{gao2017tall, krishna2017dense} is a vision–language task that aims to localise the snippet that corresponds to a given natural language query with start and end timestamps in a video.
Existing methods fall into two main categories: proposal-based approaches~\cite{gao2017tall, anne2017localizing, xu2019multilevel, zhang2020learning, wang2022negative, hou2023cone}, which generate candidate segments before matching them with the query; and proposal-free approaches~\cite{xu2019multilevel, yuan2019find, zhang2020span, mun2020local, lei2021detecting, moon2023query, yan2023unloc}, which directly predict the temporal boundaries in an end-to-end manner.
Recently, VLM-based methods~\cite{ren2024timechat, huang2024vtimellm, li2024groundinggpt, guo2024trace, zeng2024timesuite, liu2024bench, guo2025vtg} have shown competitive performance by leveraging generalised knowledge from VLMs pre-training, while also maintaining the conversational capabilities of language models.
However, both traditional and VLM-based approaches rely heavily on large amounts of labelled data and struggle to generalise to unseen domains~\cite{yuan2021closer, li2022compositional, wang2024routing, fang2024not, cheng2024shine}.
Although some work has explored cross-dataset generalisation~\cite{liu2024unsupervised, wang2025timezero, li2025temporalr1}, this setting remains challenging due to two key issues:
(i) target domain videos are often large, incurring high storage and adaptation costs; and (ii) temporal boundary annotation is labour-intensive.
Moreover, our experiments show that the recent works, e.g., TimeZero~\cite{wang2025timezero} and Temporal-R1~\cite{li2025temporalr1}, still perform poorly in cross-domain generalisation despite utilising GPRO to improve the model's generalisability with reinforcement learning.
To address this, we propose a data-efficient cross-domain temporal grounding setting and a method that achieves performance comparable to fully supervised training using only a small number of unlabelled target domain videos.
This significantly reduces resource requirements while improving cross-domain generalisability.

\textbf{Uncertainty Estimation.}
Uncertainty estimation aims to capture either data-inherent noise (aleatoric uncertainty) or model-driven ambiguity (epistemic uncertainty)~\cite{der2009aleatory,kendall2017uncertainties}. In vision tasks, epistemic uncertainty is commonly approximated via Bayesian neural networks such as MC Dropout~\cite{2015Dropout,hu2022learning}, or through approximate reasoning methods~\cite{2020Transferable, 2017Implicit}. For VLMs, existing approaches quantify uncertainty through logit entropy~\cite{guo2017calibration}, verbalized confidence enhanced by CoT prompting~\cite{xiong2023can}, or consistency-based diagnostics~\cite{zhao2023verify}. Recent work also explores semantic-level uncertainty by clustering outputs and computing entropy in embedding space~\cite{farquhar2024detecting}.
In contrast, our URPA estimates epistemic uncertainty by computing the standard deviation across multiple rollout predictions, offering a lightweight and task-aligned signal to guide self-learning in unlabelled target domains.

\begin{figure*}[h!]
   \centering 
   \vspace{-8pt}
   \hspace{-1mm}\includegraphics[width=14.2cm]{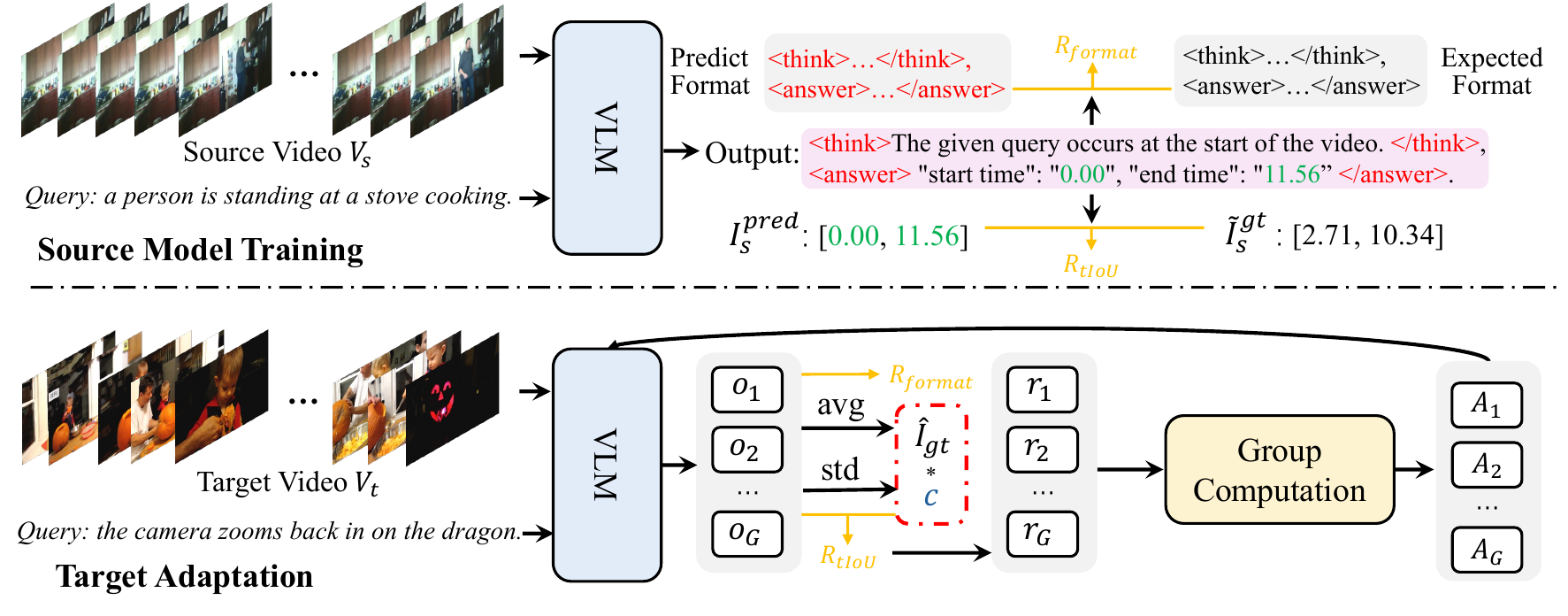}
   \vspace{-15pt}
   \caption{Uncertainty-quantified Rollout Policy Adaptation (URPA): During source model training, we perform supervised GRPO training using labelled videos $V_s$. Specifically, the format reward $R_{\text{format}}$ encourages the model to ``think first and then answer,'' while accuracy reward $R_{\text{tiou}}$ aligns the predicted temporal grounding $I_s^{\text{pred}}$ with the relaxed ground truth $\ \tilde{I}_s^{\text{gt}}$ for supervised learning. 
In target model knowledge adaptation, we adapt the model using $K$ unlabelled target videos. For each video $V_t$, we first compute the average output over $G$ rollouts to obtain a pseudo label $\hat{I}_t^{\text{gt}}$. Then calculate the standard
deviation across these rollouts and transform it into a confidence score $c$ to quantify uncertainty on pseudo labels, which is then used to weight different pseudo-labels when constructing a weighted reward function for test-time target model adaptation.
}\label{fig:framework}
\vspace{-10pt}
\end{figure*}

\vspace{-5pt}
\section{Uncertainty-quantified Rollout Policy Adaptation}
\vspace{-5pt}
\subsection{Problem Definition}
\vspace{-5pt}
In training a model for a new target domain video temporal grounding, we consider a labelled source domain $D_s=\{V_s^i,
I_s^i\}_{i=1}^{N}$ and an target domain with a small number of unlabelled videos $D_t=\{V_t^i\}_{i=1}^K$, where $I_s^i$ is the annotated interval for $V_s^i$, the number of training videos in the source and target
domains are $N$ and $K$ respectively, with $K \ll N$.   %
A distribution shift exists between the source and target domains, i.e. $\mathcal{P}_s \neq \mathcal{P}_t$.  
The model is first pre-trained on the labelled source domain to learn some general knowledge of video temporal grounding. It then learns to adapt at test-time without labelled training in the target domain
through a data-efficient unlabelled target domain videos. This real-time cross-domain unlabelled adaptation approach to video temporal grounding aims to optimise cross-domain knowledge transfer with only a few target videos without any annotations.

\vspace{-5pt}
\subsection{Remark on Group Relative Policy Optimization}
\vspace{-5pt}

In reinforcement learning, Group Relative Policy Optimization (GRPO) is a variant of Proximal Policy Optimization (PPO) that eliminates the need for a critic function. Instead, it directly evaluates the quality of predictions using a group of sampled responses. 
Given a question $q$, the model generates $G$ candidate rollout responses $o = \{o_1, o_2, \ldots, o_G\}$ through policy sampling. A reward function then computes scores $r = \{r_1, \ldots, r_G\}$ by comparing each candidate with the ground truth.
To normalize these rewards, GRPO computes their mean and standard deviation. The quality of each response is estimated by:
\begin{equation}
A_i = \frac{r_i - \text{mean}(r)}{\text{std}(r)},
\end{equation}
where $A_i$ denotes the normalized advantage of the $i$-th response.
GRPO optimizes the policy $\pi_{\theta}$ to maximize $A_i$, thereby encouraging beneficial deviation from the initial policy. A KL-divergence regularization term is further incorporated to constrain excessive deviation:
\begin{small}
\begin{equation}
\max_{\pi_\theta} \mathbb{E}_{o \sim \pi_{\theta_{\text{old}}}(p)} \left[
\left( \sum_{i=1}^{G} \frac{\pi_\theta(o_i)}{\pi_{\theta_{\text{old}}}(o_i)} \cdot A_i \right)
- \beta \, D_{\mathrm{KL}}\left( \pi_\theta \, \| \, \pi_{\text{ref}} \right)
\right],
\end{equation}
\end{small}
where $\beta$ is a regularization coefficient controlling the strength of the penalty.
Although GRPO has shown strong performance on math reasoning tasks, it struggles to generalise under domain shifts, especially for temporal grounding on unlabelled videos. This is due to both the difficulty of label-free optimisation and the high storage and adaptation cost associated with large-scale video data. 
To overcome these limitations, we introduce a data-efficient cross-domain temporal grounding setting, which consists of source domain model training and lightweight target domain adaptation.

\vspace{-5pt}
\subsection{Source Model Training}
\vspace{-5pt}
VLMs focus on coarse-grained understanding but lack fine-grained temporal localization capability. Therefore, post-training is necessary to equip the model with temporal reasoning and grounding abilities.
To this end, we pursue two main objectives. First, we aim to stimulate explicit reasoning chains within the model, enabling it to perform structured temporal inference. Second, we design task-specific reward functions to guide the model towards learning accurate temporal grounding, encouraging outputs that precisely align with relevant video segments.

\textbf{Format Reward.} 
To facilitate explicit reasoning, following \cite{wang2025timezero}, we introduce a format reward that encourages the model to structure its outputs according to a specified template. Specifically, we require the model to present its thought process within \texttt{<think>...</think>} tags, and its final answer within \texttt{<answer>...</answer>} tags. We use regular expression matching to determine whether the model's output adheres to the required format as follows:
\vspace{-5pt}
\begin{equation}
R_{\text{format}} = 
\begin{cases}
1, & \text{if output matches format,} \\
0, & \text{if output doesn't match format.}
\end{cases}
\end{equation}

\vspace{-5pt}
\textbf{Accuracy Reward.}
A core objective for source training is to design a well-crafted reward that guides the model to perform reasoning based on reward variations, thereby improving the quality of temporal grounding predictions. To guide the model towards this goal, we design an accuracy reward based on the overlap between the predicted and ground-truth temporal intervals.

However, event boundaries in videos are inherently ambiguous, and ground-truth annotations across datasets often exhibit labelling biases due to human subjectivity. Such biases can lead to performance degradation when models trained on the source domain are applied to the target domain.
To mitigate this issue, we first relax the ground-truth interval $I_{\text{s}} = [{\tau}_s^{\text{start}}, {\tau}_s^{\text{end}}]$ by extending both boundaries by a fixed proportion $\alpha$ of the event duration, yielding the relaxed ground-truth $\tilde{I}_{\text{s}}^{gt}$:
\begin{equation}
\tilde{I}_{\text{s}}^{gt} = 
\left[
\tilde{\tau}_s^{\text{start}},\; \tilde{\tau}_s^{\text{end}}
\right]
=
\left[
\max(0, {\tau}_s^{\text{start}} - \alpha ({\tau}_s^{\text{end}} - {\tau}_s^{\text{start}})),\;
\min(1, {\tau}_s^{\text{end}} + \alpha ({\tau}_s^{\text{end}} - {\tau}_s^{\text{start}})
\right],
\label{eq:equ_relax}
\end{equation}
where ${\tau}_s^{\text{start}}$ and ${\tau}_s^{\text{end}}$ are normalized timestamps within $[0,1]$, and $\alpha$ is set to 0.1. 
Let $I_s^{\text{pred}} = [p_s^{\text{start}}, p_s^{\text{end}}]$ denote the predicted temporal interval.  
We then compute the relaxed temporal Intersection over Union (tIoU) between $I_s^{\text{pred}}$ and $\tilde{I}_s^{\text{gt}}$ as:
\begin{equation}
R_{\text{tIoU}} = \frac{|\tilde{I}_s^{\text{gt}} \cap I_s^{\text{pred}}|}{|\tilde{I}_s^{\text{gt}} \cup I_s^{\text{pred}}|} = \frac{\max(0, \min(p_s^{\text{end}}, \tilde{\tau}_s^{\text{end}}) - \max(p_s^{\text{start}}, \tilde{{\tau}}_s^{\text{start}}))}{\max(p_s^{\text{end}}, \tilde{\tau}_s^{\text{end}}) - \min(p_s^{\text{start}}, \tilde{\tau}_s^{\text{start}})},
\end{equation}
where $\tilde{\tau}_s^{\text{start}}$ and $\tilde{\tau}_s^{\text{end}}$ are the relaxed ground-truth boundaries defined in Eq.(\ref{eq:equ_relax}). %
A higher $R_{\text{tIoU}}$ indicates better alignment between the predicted and relaxed ground-truth segments.
Thus, the overall supervised reward function for source domain training is defined as:
\begin{equation}
R_s = 0.5 \times R_{\text{format}} + 0.5 \times R_{\text{tIoU}},
\end{equation}
Supervised training on the source domain equips the model with basic spatio-temporal reasoning capability, but it still suffers from performance degradation on the target domain due to the domain shift between source and target domains, making target unsupervised adaptation essential.

\vspace{-5pt}
\subsection{Target Adaptation}
\vspace{-5pt}
GRPO requires ground-truth labels to compute rewards, which are unavailable in the target domain. Meanwhile, traditional domain adaptation methods typically assume full access to target domain data during training. Nevertheless, this assumption becomes impractical for large-scale video tasks due to substantial storage and computational costs.
Thus we consider a more realistic setting where only a small number of $K$ unlabelled target videos are available for adaptation, with $K \ll N$.

While pretraining on the source domain provides the model with general temporal grounding ability, performance still degrades under domain shift. To bridge this gap, we leverage the multiple candidate responses generated during GRPO optimization. For each target sample, the policy model samples $G$ rollout response candidates. Although each individual response may be noisy, their aggregated statistics offer a more stable approximation. We therefore construct pseudo temporal labels by averaging predicted start and end timestamps:
\begin{equation}
\hat{\tau}_t^{\text{start}} = \frac{1}{G} \sum_{j=1}^{G} (p_t^{\text{start}})_{(j)}, \quad 
\hat{\tau}_t^{\text{end}} = \frac{1}{G} \sum_{j=1}^{G} (p_t^{\text{end}})_{(j)},
\end{equation}
where $(p_t^{\text{start}})_{(j)}$ and $(p_t^{\text{end}})_{(j)}$ denote the predicted timestamps from the $j$-th sampled response. These pseudo labels act as soft supervision after processed with Eq.(\ref{eq:equ_relax}) for test-time adaptation.
However, not all pseudo labels are equally reliable. To evaluate their quality, we estimate prediction uncertainty. Inspired by Bayesian deep learning, we treat the $G$ rollout responses as approximate samples from a predictive distribution, analogous to the Monte Carlo Dropout approach. This allows us to use the standard deviation of the predicted timestamps as a proxy for uncertainty:
\begin{equation}
u = \sigma \left(\left\{(p_t^{\text{start}})_{(j)}\right\}_{j=1}^{G}\right) + 
    \sigma\left(\left\{(p_t^{\text{end}})_{(j)}\right\}_{j=1}^{G}\right),
\end{equation}

The uncertainty $u$ is then converted into a confidence score $c$ via an exponential decay function:
\begin{equation}
c = \exp(-\gamma u),
\label{eq:confidence}
\end{equation}
where $\gamma$ is a hyperparameter that controls the influence of uncertainty on confidence.
Finally, this confidence score is used to weight the contribution of the pseudo labels during reward computation. Specifically, we scale the temporal grounding reward (\( R_{\text{tIoU}} \)) by the confidence, while keeping the format reward unweighted:
\begin{equation}
\label{eq:confidence_formulation}
R_s = 0.5 \times R_{\text{format}} + 0.5 \times R_{\text{tIoU}} \times c.
\end{equation}

This confidence-aware reward formulation allows the model to better exploit informative pseudo labels while mitigating the effects of noisy or uncertain predictions, leading to more robust test-time adaptation in the absence of ground-truth labels.

\vspace{-8pt}
\section{Theoretical Analysis}
\vspace{-7pt}
\label{sec:theory}

This section proves that the empirical standard deviation obtained from multiple GRPO rollouts converges to the Bayesian predictive standard deviation, thereby quantifying epistemic uncertainty.

\begin{theorem}
\label{thm:rollout_std}
Fix an input $x$.  
Let $\pi_\theta(\tau\mid x)$ be the rollout policy learned by GRPO and
let $p^\star(\tau\mid x)$ denote the Bayesian predictive distribution
of the same model class trained on the source data.
Assume
\textbf{(i)} rollouts drawn from $\pi_\theta$ are i.i.d.;  
\textbf{(ii)} there exists a constant $M<\infty$ such that  
\(
\mathbb E_{\pi_\theta}[\tau^{2}]\le M
\) and
\(
\mathbb E_{p^\star}[\tau^{2}]\le M;
\)
\textbf{(iii)} the Kullback–Leibler divergence is finite,
\(
\varepsilon=\mathrm{KL}\!\bigl(\pi_\theta\,\|\,p^\star\bigr)<\infty .
\)
Draw $G$ independent rollouts
$\{\tau^{(i)}\}_{i=1}^{G}\sim\pi_\theta(\cdot\mid x)$ and define
{\footnotesize
\[
\overline{\tau}_G = \frac1G\sum_{i=1}^{G} \tau^{(i)},
\qquad
\widehat{\sigma}_G(x) =
\sqrt{\frac1G\sum_{i=1}^{G}\bigl(\tau^{(i)}-\overline{\tau}_G\bigr)^{2}}.
\]}
As $G\!\to\!\infty$ and GRPO training drives
$\varepsilon\!\to\!0$, the estimator
$\widehat{\sigma}_G(x)$ converges to the Bayesian predictive standard deviation 
$\operatorname{Std}_{p^\star}[\tau\mid x]$, providing a consistent measure of epistemic uncertainty.
\end{theorem}

\begin{proof}
For clarity we give the main steps; full details are in the supplementary material Sec.~\ref{sec:rollout_varPdetail_n}.

\textbf{Step 1: Consistency under the GRPO policy.}
Let $\mu_\pi=\mathbb E_{\pi_\theta}[\tau]$ and
$\sigma_\pi(x)=\operatorname{Std}_{\pi_\theta}[\tau\mid x]$.
Because the $G$ rollouts $\{\tau^{(i)}\}$ are i.i.d.\ and
$\mathbb E_{\pi_\theta}[\tau^{2}]\!\le\!M$ (Theorem assumption (ii)),
the weak law of large numbers yields
\begin{small}
\begin{equation}\label{eq:wlln-main}
\overline{\tau}_G\xrightarrow[]{p}\mu_\pi,
\qquad
\widehat{\sigma}_G(x)\xrightarrow[]{p}\sigma_\pi(x)\quad(G\!\to\!\infty).
\end{equation}
\end{small}

\textbf{Step 2: Relating the GRPO and Bayesian standard deviations.}
With $\varepsilon=\mathrm{KL}(\pi_\theta\|p^\star)$ (assumption (iii)),
Pinsker’s inequality gives a total-variation bound
\begin{equation}\label{eq:pinsker-main}
\|\pi_\theta-p^\star\|_{\mathrm{TV}}\le\sqrt{\tfrac12\,\varepsilon}.
\end{equation}
Using \eqref{eq:pinsker-main}, the bounded-moment assumption and
$\operatorname{Var}[\tau]=\mathbb E[\tau^{2}]-(\mathbb E[\tau])^{2}$
one obtains the variance gap bound:
\begin{small}
\begin{equation}\label{eq:var-gap-main}
\bigl|\sigma_\pi^2(x)-\sigma_\star^2(x)\bigr|
      \le 4M\sqrt{\varepsilon},
\qquad
\sigma_\star^2(x)=\operatorname{Var}_{p^\star}[\tau\mid x].
\end{equation}
\end{small}
Applying the identity $|a - b| \le \sqrt{|a^2 - b^2|}$ for non-negative $a$, $b$, we get:
\begin{small}
\begin{equation}
\bigl|\sigma_\pi(x)-\sigma_\star(x)\bigr|
\;\le\; \sqrt{4M\sqrt{\varepsilon}}.
\end{equation}
\end{small}

\textbf{Step 3: Convergence to Bayesian predictive standard deviation.}
GRPO optimisation decreases $\varepsilon$ over training, so
$\sqrt{\varepsilon}\!\to\!0$. Combining
\eqref{eq:wlln-main} and the above yields
\begin{small}
\begin{equation}
\bigl|\widehat{\sigma}_G(x)-\sigma_\star(x)\bigr|
\;\le\;
\underbrace{\bigl|\widehat{\sigma}_G(x)-\sigma_\pi(x)\bigr|}_{\text{vanishes by \eqref{eq:wlln-main}}}
+
\underbrace{\bigl|\sigma_\pi(x)-\sigma_\star(x)\bigr|}_{\text{vanishes as } \varepsilon \to 0},
\end{equation}
\end{small}
which tends to zero in probability as
$G\!\to\!\infty$ and $\varepsilon\!\to\!0$.
Hence
\begin{small}
\begin{equation}
\widehat{\sigma}_G(x)\xrightarrow[]{p}\sigma_\star(x),
\end{equation}
\end{small}
showing that the rollout standard deviation is a consistent estimator of
epistemic uncertainty.
\end{proof}

\begin{table*}[t]
\centering
\small
\caption{Performance comparisons on three cross-domain temporal grounding benchmarks. The best is in \textbf{bold}.}
\vspace{-5pt}
\setlength{\tabcolsep}{4pt}
\begin{subtable}[t]{\textwidth}
\centering
\resizebox{1.0\textwidth}{!}{
\begin{tabular}{l|c|c|c|c|c|c}
\hline
\multirow{2}{*}{Method} & \multicolumn{2}{c|}{Charades$\rightarrow$ActivityNet} & \multicolumn{2}{c|}{ActivityNet$\rightarrow$TACoS} & \multicolumn{2}{c}{TACoS$\rightarrow$Charades} \\\cline{2-7}
& R@0.5 & R@0.7 &  R@0.3 & R@0.5 & R@0.5 & R@0.7  \\
\hline
\multicolumn{7}{c}{Full-dataset Unsupervised Domain Adaptation} \\ \hline
CBP~\cite{wang2020temporally}     & 27.46 & 15.37 &  25.33 & 21.79  & 22.38 & 11.95  \\
SCDM~\cite{yuan2019semantic}    & 28.02 & 15.84  & 22.68 & 17.45 &  35.95 & 25.18  \\
CMIN~\cite{zhang2019cross}    & 34.25 & 18.63 & 20.51 & 15.04  & 28.06 & 18.22 \\
CSMGAN~\cite{lin2020weakly}  & 36.92 & 20.04 & 29.63 & 18.07 & 36.45 & 22.86 \\
2DTAN~\cite{zhang2020learning}   & 39.17 & 21.76 & 33.72 & 21.16 & 25.81 & 17.37 \\
DRN~\cite{zeng2020dense}     & 41.39 & 24.27 & 32.07 & 19.96 & 36.16 & 24.52 \\
MMN~\cite{wang2022negative}     & 44.06 & 24.98 & 36.94 & 22.08 & 33.73 & 20.04 \\
UDA-TSL~\cite{liu2024unsupervised} & \textbf{49.48} & \textbf{32.15} & \textbf{42.40} & \textbf{29.83} & \textbf{41.39} & \textbf{28.63} \\ \hline
% \multicolumn{7}{c}{Target Data-Efficient Supervised Learning} \\ \hline
% Target Supervised GRPO (200-shot) & 42.75 & 22.42 & 22.20 & 11.17 & 59.38 & 33.36\\
% \hline
\multicolumn{7}{c}{Domain Generalisation} \\ \hline
Qwen2.5-7B~\cite{yang2024qwen2} & 15.18& 7.90&7.70 & 2.77& 32.93 & 15.35 \\
+ GRPO (Source Training only) & 35.61 & 16.61 & 20.80 & 9.87 & 53.51 & 28.69\\
\rowcolor{purple!10} + URPA (Source Training only) & 36.46 & 18.78 & 21.58 & 10.26 & 54.40 & 29.81\\
\hline
\multicolumn{7}{c}{Data-Efficient Unsupervised Domain Adaptation} \\ \hline
\rowcolor{purple!10} URPA (with 100-shot Target Adaptation) & 40.13 & 20.88 & 21.82 & 10.16 & 54.78 & 30.08 \\
\rowcolor{purple!10} URPA (with 200-shot Target Adaptation) & \textbf{42.57} & \textbf{21.25} & \textbf{21.97} & \textbf{10.38} & \textbf{55.54}& \textbf{32.04}\\
\hline
\end{tabular}}
\end{subtable}

\vspace{0.5em}

\setlength{\tabcolsep}{4pt}
\begin{subtable}[t]{\textwidth}
\centering
\resizebox{1.0\textwidth}{!}{
\begin{tabular}{l|c|c|c|c|c|c}
\hline
\multirow{2}{*}{Method} & \multicolumn{2}{c|}{TACoS$\rightarrow$ActivityNet} & \multicolumn{2}{c|}{Charades$\rightarrow$TACoS} & \multicolumn{2}{c}{ActivityNet$\rightarrow$Charades} \\\cline{2-7}
& R@0.5 & R@0.7 & R@0.3 & R@0.5 & R@0.5 & R@0.7 \\\hline
\multicolumn{7}{c}{Full-dataset Unsupervised Domain Adaptation} \\ \hline
CBP~\cite{wang2020temporally}     & 18.94 & 11.93 & 22.88 & 19.26 & 32.82 & 14.39 \\
SCDM~\cite{yuan2019semantic}    & 19.65 & 11.80 & 18.97 & 16.82 & 52.56 & 34.82  \\
CMIN~\cite{zhang2019cross}    & 22.17 & 13.72 & 19.38 & 15.34 & 45.03 & 31.74  \\
CSMGAN~\cite{lin2020weakly}  & 23.88 & 14.67  & 25.43 & 16.12 & 45.60 & 32.28 \\
2DTAN~\cite{zhang2020learning}   & 24.90 & 16.38 & 30.12 & 19.81 & 36.34 & 22.61 \\
DRN~\cite{zeng2020dense}     & 24.93 & 18.52 & 28.60 & 16.73 & 50.47 & 29.02 \\
MMN~\cite{wang2022negative}     & 28.29 & 20.86 & 34.09 & 19.17 & 50.78 & 23.17 \\
UDA-TSL~\cite{liu2024unsupervised} & \textbf{33.54} & \textbf{26.16} & \textbf{36.42} & \textbf{25.48} & \textbf{60.26} & \textbf{41.03} \\
% \hline
% \multicolumn{7}{c}{Target Data-Efficient Supervised Learning} \\ \hline
% Target Supervised GRPO (200-shot) & 44.78 & 24.36 & 18.83 & 9.65 & 66.77 & 41.56\\ 
\hline
\multicolumn{7}{c}{Domain Generalisation} \\ \hline
Qwen2.5-7B~\cite{yang2024qwen2} &15.18&7.90&7.70& 2.77&32.93 & 15.35\\
 + GRPO (Source Training only) & 34.82 & 17.02 & 12.55 & 5.13 & 62.13 & 36.06\\
\rowcolor{purple!10}  + URPA (Source Training only) & 36.23&18.13&13.91&6.27&63.36 & 37.47 \\ 
\hline
\multicolumn{7}{c}{Data-Efficient Unsupervised Domain Adaptation} \\ \hline
\rowcolor{purple!10} URPA (with 100-shot Target Adaptation) &38.41 &19.80&14.49&7.23 &64.35 &38.76 \\
\rowcolor{purple!10} URPA (with 200-shot Target Adaptation) &\textbf{41.83} & \textbf{21.84} & \textbf{16.62} & \textbf{8.25} & \textbf{65.12} & \textbf{39.57}\\
\hline
\end{tabular}}
\end{subtable}
\vspace{-10pt}
\label{tab:main_results}
\end{table*}

\vspace{-10pt}
\section{Experiments}
\label{sec:experiment}
To evaluate the effectiveness of the proposed method, we conduct experiments on three widely used temporal grounding datasets: TACoS~\cite{regneri2013grounding}, ActivityNet Captions~\cite{caba2015activitynet}, and Charades-STA~\cite{sigurdsson2016hollywood}.

\subsection{Experimental Setup} 

\textbf{Datasets.} We evaluate our method on three benchmark datasets: TACoS, ActivityNet Captions, and Charades-STA.
ActivityNet Captions contains approximately 20000 untrimmed YouTube videos annotated with 100000 natural language descriptions. Following the standard split, we use 37417 sentence-video pairs for training, and 17031 for testing.
TACoS consists of 127 cooking-related videos. We adopt the public split, which includes 10146 and 4083 query-segment pairs for training and testing, respectively.
Charades-STA is built on the Charades dataset, containing 12408 training and 3720 testing moment-query pairs. Codes are given in supplemental materials.
%
% To evaluate cross-domain generalisation and adaptation, we follow the conventional setup for cross-domain temporal grounding. Specifically, for each experiment, we treat one dataset as the source domain (with labelled training data) and the other two as target domains (unlabelled), resulting in six experimental configurations.

\textbf{Baselines.}
To evaluate both cross-domain generalisation and adaptation, we follow the standard protocol for temporal grounding under domain shift. In each experiment, one dataset serves as the labelled source domain, while the remaining two act as unlabelled target domains, yielding six cross-domain configurations.
We compare our method under the data-efficient unsupervised adaptation setting, where the source-pretrained model is adapted using only K = 100 or 200 unlabelled target videos with our URPA. This setting is contrasted with the following baselines: (1) Full-set adaptation: The model is trained on the labelled source domain and adapted using the entire unlabelled target domain. We report results for several state-of-the-art unsupervised adaptation methods, including CBP~\cite{wang2020temporally}, SCDM~\cite{yuan2019semantic}, CMIN~\cite{zhang2019cross}, CSMGAN~\cite{lin2020weakly}, 2DTAN~\cite{zhang2020learning}, DRN~\cite{zeng2020dense}, MMN~\cite{wang2022negative}, and UDA-TSL~\cite{liu2024unsupervised}.
(2) Domain Generalisation: We evaluate the base Qwen2.5-7B model~\cite{yang2024qwen2} (without any fine-tuning), a GRPO-trained model on the source domain, and a model trained using only the source training phase of our URPA framework. All three are trained exclusively on the source domain and directly evaluated on the target domain without any access to target data.
% % 
% For our data-efficient unsupervised adaptation setting, we report performance after adapting a source-trained model using 100 or 200 unlabelled target videos.
% For the domain generalisation setting, we evaluate both the vanilla Qwen2.5-7B model~\cite{yang2024qwen2} and its GRPO-adapted variant, trained on the source domain and directly tested on the target domain without adaptation.

\textbf{Implementation Details.}
We first fine-tune Qwen2.5-7B on the labelled source domain, and then perform data-efficient adaptation on 100 or 200 randomly selected samples from the unlabelled target domain. In all experiments, the maximum prompt length is set to 4096, the maximum response length to 2048, the number of rollouts to 8, batch size to 16, and we train for 1 epoch. All models are implemented in PyTorch on 32 NVIDIA V100 GPUs. Codes are in supplemental materials.

\subsection{Results and Analysis}
\begin{table}[t]
\centering
% \small
\caption{Ablation study on Charades $\rightarrow$ Activitynet dataset showing the impact of uncertainty estimation and soft labelling.}
\setlength{\tabcolsep}{4pt}
\resizebox{1.0\textwidth}{!}{
\begin{tabular}{l|c|c|c|c}
\hline
Variant & {R@0.3} & {R@0.5} & {R@0.7} & {mIoU} \\
\hline
Qwen2.5-7B & 25.11 & 15.18 & 7.90 & 18.47 \\
Qwen2.5-7B + Our GRPA Target Adaptation (100-shot) & 2.67 & 1.46 & 0.68 & 1.96 \\
Qwen2.5-7B + Source Training with GRPO & 53.91 & 35.61 & 16.61 & 35.81 \\
Qwen2.5-7B + Source Training with our GRPA & 55.23 & 36.46 & 18.78 & 37.64 \\
\hline
URPA w/o uncertainty-quantification + w/o relaxed tIOU (200-shot) & 58.58 & 31.15 & 20.75 & 39.86 \\
URPA w/o uncertainty + w/ relaxed tIOU (200-shot)& 63.72 & 41.48 & 20.47 & 43.10 \\
% Target Supervised Learning (200-shot) & 62.51 & 42.74 & 22.42 & 42.56 \\
\hline
\rowcolor{purple!10} URPA (200-shot) & \textbf{64.61} & \textbf{42.57} & \textbf{21.25} & \textbf{43.65} \\
\hline
\end{tabular}
}
\label{tab:ablation_charades}
\end{table}
\begin{table*}[t]
\centering
\caption{Ablation study of GRPO hyperparameters on the Charades $\rightarrow$ ActivityNet task: (a) effect of reward scaling factor $\gamma$, and (b) effect of rollout count $G$ during adaptation.}
\small
\vspace{-3pt}
\hspace{-7mm}
\renewcommand{\arraystretch}{1.1}
\setlength{\tabcolsep}{7pt} 
\begin{subtable}[t]{0.47\textwidth}
\centering
\begin{tabular}{c|c|c|c|c}
\hline
$\gamma$ & R@0.3 & R@0.5 & R@0.7 & mIoU \\
\hline
2  & 59.01 & 40.04 & 19.98 & 40.09 \\
5  & 59.06 & \textbf{40.44} & 20.13 & 40.25 \\
\cellcolor{purple!10}10 & \cellcolor{purple!10}\textbf{59.12} & \cellcolor{purple!10}40.13 & \cellcolor{purple!10}\textbf{20.88} & \cellcolor{purple!10}\textbf{40.46} \\
25 & 58.86 & 39.89 & 19.97 & 40.11 \\
\hline
\end{tabular}
\caption{Varying $\gamma$ with 100-shot unsupervised adaptation.}
\label{tab:gamma}
\end{subtable}
\hspace{-2mm}
\renewcommand{\arraystretch}{0.9}
\begin{subtable}[t]{0.5\textwidth}
\centering
\begin{tabular}{c|c|c|c|c}
\hline
$G$ (rollouts) & R@0.3 & R@0.5 & R@0.7 & mIoU \\
\hline
4 & 63.41 & 42.27 & 21.39 & 42.55 \\
\cellcolor{purple!10}8   & \cellcolor{purple!10}\textbf{64.61} & \cellcolor{purple!10}\textbf{42.57} & \cellcolor{purple!10}21.25 & \cellcolor{purple!10}\textbf{43.65} \\
16  & 63.19 & 41.78 & \textbf{21.73} & 42.40 \\
32  & 62.56 & 41.65 & 21.54 & 42.15 \\
64  & 60.08 & 40.98 & 20.86 & 41.37 \\
\hline
\end{tabular}
\caption{Varying $G$ with 200-shot unsupervised adaptation.}
\label{tab:rollout}
\end{subtable}
\vspace{-15pt}
\label{tab:hyperparam_ablation}
\end{table*}

\textbf{Experimental Results.}  
Tab.~\ref{tab:main_results} shows the performance of our method on the unsupervised cross-domain temporal grounding tasks. We report results using our data-efficient unsupervised adaptation approach with \(K = 100\) and \(K = 200\) unlabelled target samples. We compare against several baselines, including: (1) traditional UDA methods trained on full labelled source and unlabelled target domain data; (2) zero-shot domain generalisation methods trained only on the source domain and the base Qwen2.5-7B model.
Our results demonstrate that data-efficient adaptation with just a small number of unlabelled target samples can significantly improve the performance of the source-pretrained model. Compared to full-set UDA methods, our method achieves competitive or even better performance in several settings. In particular, for TACoS $\rightarrow$ ActivityNet, TACoS $\rightarrow$ Charades, and ActivityNet $\rightarrow$ Charades tasks, our method outperforms state-of-the-art UDA methods in R@0.5. These results highlight the effectiveness and efficiency of our approach in realistic cross-domain scenarios with limited adaptation data. We also compare our approach with data-efficient target-supervised learning baselines in Appendix~\ref{sec:more}, further validating its effectiveness.

\textbf{Ablation Study.}
In Tab.~\ref{tab:ablation_charades}, we present an ablation study on the Charades$\rightarrow$ActivityNet cross-domain temporal grounding task to assess the impact of different components in our framework.
The first row reports the performance of the base model, Qwen2.5-7B~\cite{yang2024qwen2}, directly applied to the temporal grounding task on ActivityNet without any adaptation. The poor performance indicates that the base model lacks inherent temporal localization capabilities.
The second row presents the performance of Qwen2.5-7B adapted to the target domain using our method without any source domain pre-training. The subpar results suggest that source domain training provides essential task-specific knowledge necessary for effective adaptation.
Rows three and four show the results of models pre-trained on the source domain using the original GRPO algorithm and our improved method, respectively, and then directly evaluated on the target domain. Both outperform the base model significantly, with our method achieving better results than the original GRPO. This demonstrates that supervised training on the source domain aids the model in learning temporal grounding, and our soft accuracy reward approach mitigates biases introduced by manual annotations during pre-training.
The last three rows analyse different modules in our target domain adaptation strategy, starting from a source-trained model. In the fifth row, pseudo labels are used directly for adaptation without modification in Eq.(\ref{eq:equ_relax}) and uncertainty quantification in Eq.(\ref{eq:confidence_formulation}), resulting in the lowest performance. In the sixth row, we apply Eq.(\ref{eq:equ_relax}) to the pseudo labels, which significantly improves results, indicating that soft label learning helps mitigate the effects of label noise. Finally, the last row introduces our uncertainty-quantified reward weighting mechanism (Eq.\ref{eq:confidence_formulation}) with soft labelling, yielding further improvements. This confirms the effectiveness of our GRPO-based uncertainty estimation strategy in guiding test-time adaptation.

\begin{figure}
% \vspace{-5pt}
  \centering
  \vspace{-10pt}\hspace{-3mm}\includegraphics[width=1.02\textwidth]{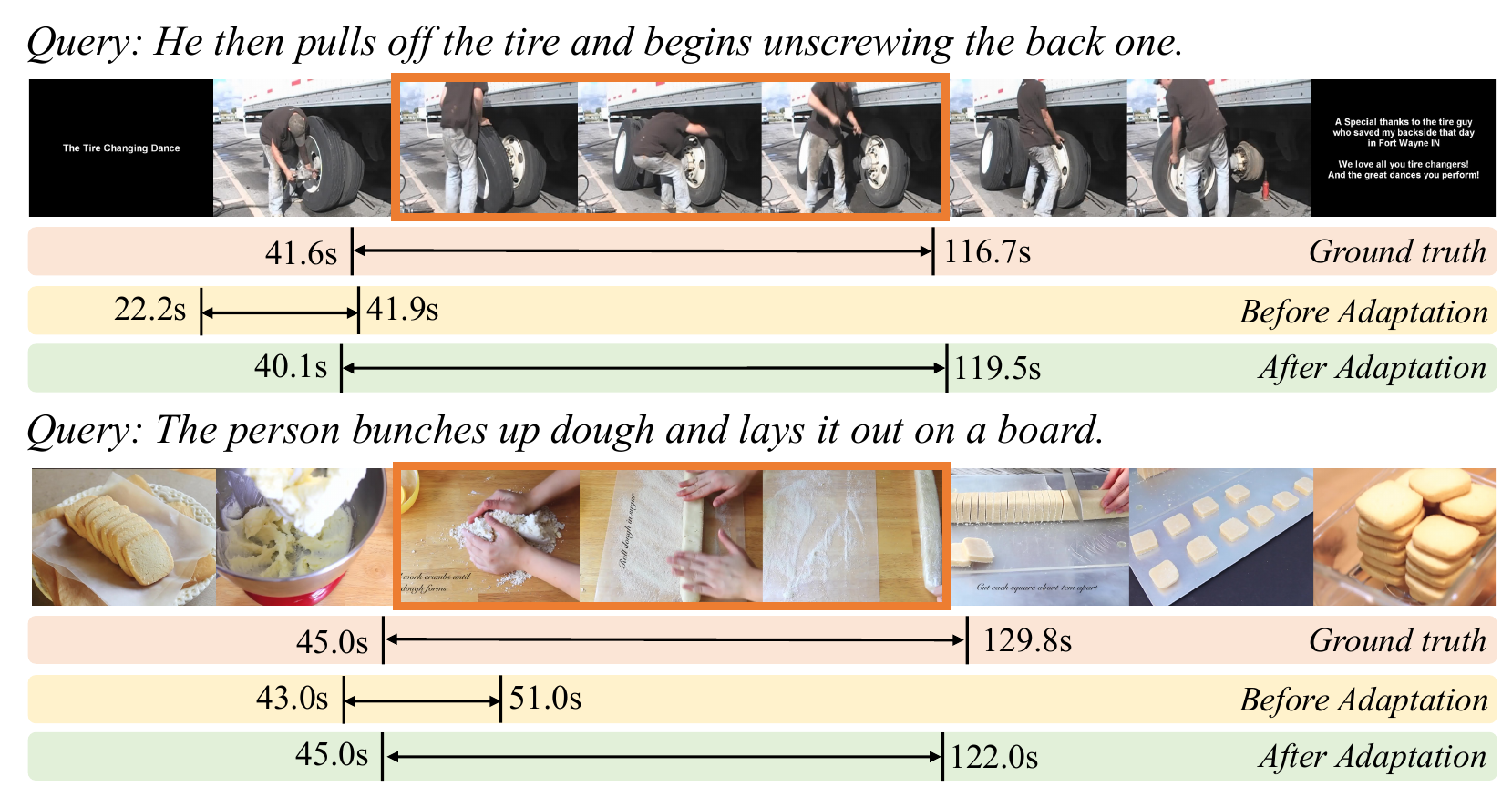}
  \vspace{-13pt}
  \caption{Qualitative Analysis on Charades $\rightarrow$ ActivityNet.}
  \label{fig:vis}
  \vspace{-10pt}
\end{figure}

\textbf{Parameter Analysis.} We analyse the effect of key hyperparameters on Charades $\rightarrow$ ActivityNet task performance. The reward scaling factor $\gamma$ (used in Eq.(\ref{eq:confidence})) controls the sensitivity to uncertainty: a larger $\gamma$ increases the influence of predicted uncertainty on reward weighting. Tab.~\ref{tab:hyperparam_ablation}(a) reports the results on the 100-shot transfer setting with varying values of $\gamma$. We observe that $\gamma = 10$ yields the best performance, but the results are relatively stable across different $\gamma$ values, indicating that our method is robust to this hyperparameter.
Tab.~\ref{tab:hyperparam_ablation}(b) investigates the impact of the number of rollouts $G$ in GRPO during 200-shot adaptation. The performance peaks when $G = 8$, and further increasing the number of rollouts does not lead to consistent gains, while incurring higher computational and memory costs. They suggest that our method is both effective and efficient under practical settings.

\textbf{Qualitative Analysis.}  Fig.~\ref{fig:vis} illustrates several qualitative examples from the Charades $\rightarrow$ ActivityNet transfer setting. Despite not using any labelled data during target adaptation, our method produces significantly more accurate temporal grounding than the baseline. Notably, in these cases where the before adaptation model completely fails to localise the correct segment, our approach is able to capture meaningful behavioural patterns in the target domain and make precise predictions. These results highlight the effectiveness of our method in adapting to unseen domains without supervision.

% \noindent\textbf{Appendix Content.}
% Effect of shot number on test-time adaptation in Appendix \ref{sec:shot_ana}, more visualisation analysis in Appendix \ref{sec: vis_more}, limitations in Appendix \ref{sec:limitation}, code information is in Appendix \ref{sec:codes}, a more detailed comparison in Appendix \ref{sec:more}, and a detailed theoretical analysis in Appendix \ref{sec:rollout_varPdetail_n}.
\vspace{-8pt}
\section{Conclusion}
\vspace{-8pt}
In this paper, we propose URPA, a data-efficient method for cross-domain temporal grounding that leverages the rollout mechanism of GRPO to estimate predictive uncertainty and assess the reliability of predictions on unlabelled target-domain videos. This allows effective test-time adaptation using only a small number of target samples.
We theoretically demonstrate that the standard deviation across rollouts approximates epistemic uncertainty. Extensive experiments across six cross-domain benchmarks validate the effectiveness of URPA in improving temporal grounding under domain shift.

\bibliography{example}
\bibliographystyle{plain}

\end{document}